\documentclass[10pt,twocolumn,letterpaper]{article}

\usepackage{cvpr}
\usepackage{booktabs}
\usepackage{multirow}
\usepackage[table]{xcolor}
\usepackage{graphicx}
\usepackage{makecell}

\usepackage{graphicx}
\usepackage{amsmath}
\usepackage{amssymb}
\usepackage{booktabs}
\usepackage{colortbl}
\usepackage{microtype}
\usepackage[section]{placeins} 

\makeatletter
\@ifpackageloaded{stfloats}{}{%
  \usepackage{stfloats} 
}
\makeatother

\usepackage[font=small,labelfont=bf]{caption}
\usepackage{subcaption} 

\numberwithin{equation}{section}

\usepackage{algorithm}
\usepackage{algpseudocode}
\makeatletter
\renewcommand{\fnum@algorithm}{\fname@algorithm~\thealgorithm}
\makeatother
\algrenewcommand\algorithmicrequire{\textbf{Inputs:}}
\algrenewcommand\algorithmicensure{\textbf{Outputs:}}
\algrenewcommand\algorithmiccomment[1]{\hfill{\(\triangleright\)\,#1}}

\usepackage{tikz}
\usepackage{xcolor}
\usepackage{fontawesome5}
\usetikzlibrary{shapes.geometric,arrows.meta,positioning,calc}

\definecolor{boxA}{HTML}{FFF8E1}
\definecolor{boxAborder}{HTML}{FFD54F}
\definecolor{boxB}{HTML}{E3F2FD}
\definecolor{boxBborder}{HTML}{64B5F6}
\definecolor{boxC}{HTML}{E8F5E9}
\definecolor{boxCborder}{HTML}{81C784}
\definecolor{mainboxfill}{HTML}{F3E5F5}
\definecolor{mainboxborder}{HTML}{BA68C8}
\definecolor{arrowcolor}{HTML}{B0BEC5}
\definecolor{titlecolor}{HTML}{4A148C}

\graphicspath{{figs/}} 

\definecolor{ijcbblue}{rgb}{0.21,0.49,0.74}
\usepackage[pagebackref,breaklinks,colorlinks,allcolors=ijcbblue]{hyperref}
\usepackage{multirow} 
\usepackage{caption}
\usepackage{cuted}

\vspace{-10em}

\title{DiffAttack: Evasion Attacks Against Face Recognition via Latent Diffusion Models}\vspace{-20pt}
\vspace{-1em}
\author{Omid Ahmadieh \qquad Nima Karimian\thanks{Corresponding author: nkarimian@usf.edu}\\
University of South Florida\\
Bellini College of Artificial Intelligence, Cybersecurity and Computing\\
Tampa, Florida, USA\\
{\tt\small oa21@usf.edu \qquad nkarimian@usf.edu}
}

\vspace{-5em}
\begin{document}
\maketitle

\begin{strip}
  \vspace{-0.5em} 
  \centering
  \includegraphics[width=0.73\linewidth]{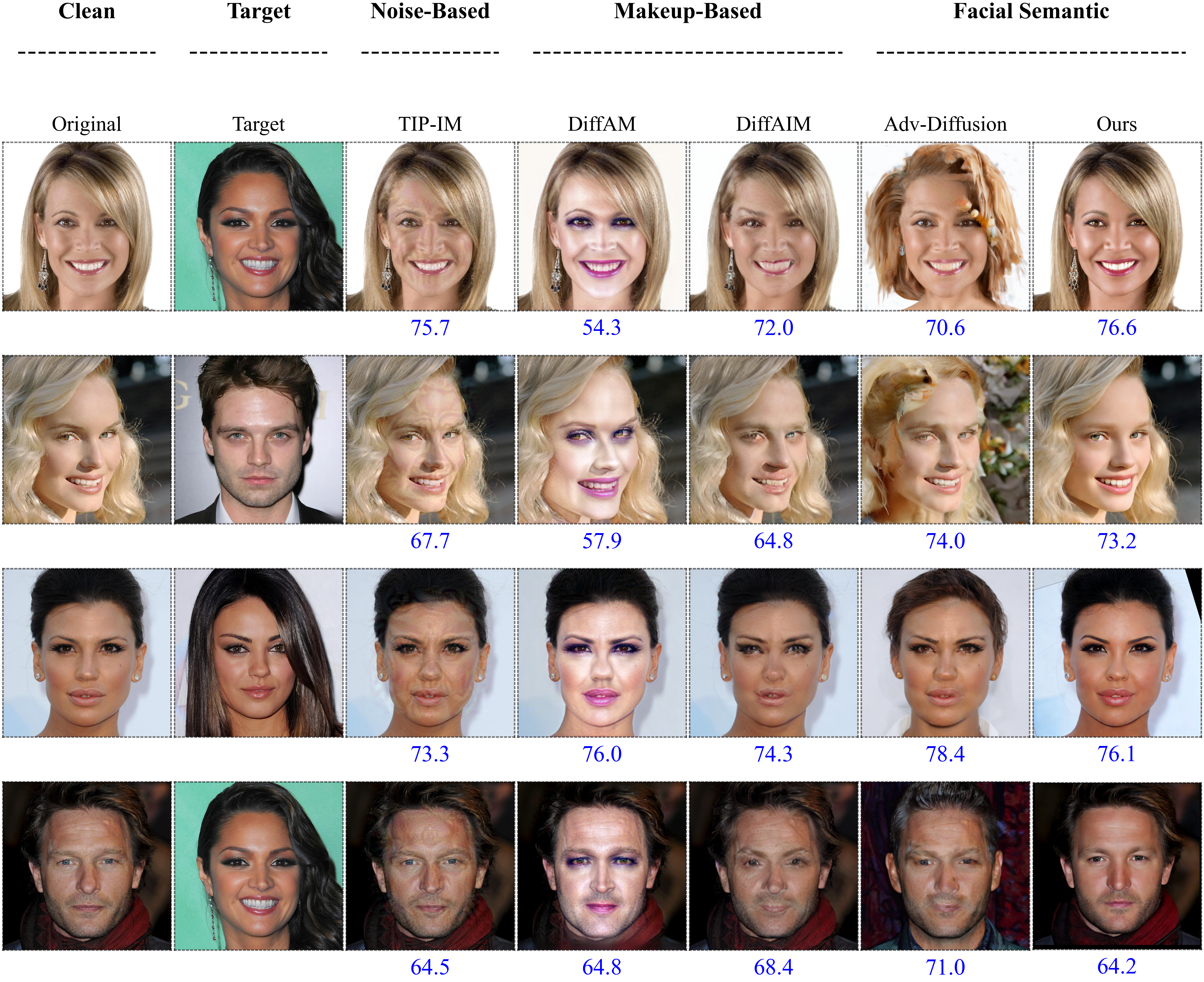}
  
  \vspace{0.5em} 
  
  \captionof{figure}{
    DiffAttack generates photorealistic, identity-aligned transformations while preserving the source's original texture, expression, and lighting. The blue values indicate Face++ confidence scores; higher scores represent stronger identity alignment with the target identity.
  }
  \label{fig:teaser}
  \vspace{-0.5em} 
\end{strip}

\begin{abstract}
Facial biometric identification relies on the distinctiveness of user attributes within a high-dimensional embedding space. However, the decision boundaries of deep face recognition (FR) systems are often sufficiently narrow that they can be conflated, rendering the models vulnerable to adversarial attacks. In such scenarios, the FR system fails to distinguish between an authentic source and a meticulously crafted adversarial face. Existing adversarial methods targeting facial biometrics are limited in both performance and their ability to generate high-quality images that are imperceptible to humans. Moreover, these methods often fail when the source and target images belong to different demographic groups or genders. To address these limitations, we present a novel approach for adversarial face generation via latent-space optimization. We leverage latent diffusion models directly to guide generation toward target identity embeddings, as measured by a face recognition model. Our proposed \textbf{DiffAttack} framework has been evaluated on standard benchmarks, such as the FFHQ and CelebA-HQ datasets. DiffAttack significantly outperforms existing adversarial techniques, achieving a high average attack success rate of 84.86\% across multiple face recognition models (e.g., FaceNet). Notably, DiffAttack demonstrates superior transferability, surpassing traditional noise-based methods by over 15.28\% and semantic-based approaches by approximately 5.21\% on benchmark datasets like FFHQ and CelebA-HQ.
\end{abstract}   
\section{Introduction}
\label{sec:intro}
Advances in deep learning have significantly improved the effectiveness of face recognition (FR) technologies, resulting in their extensive use in real-world applications such as identity authentication, surveillance systems, and social media tagging~\cite{mi2024privacy,pinto2011scaling}. While these technologies have brought substantial convenience, their large-scale adoption has also introduced serious security concerns. By leveraging massive collections of publicly available facial images, modern FR systems can infer social relationships~\cite{zhang2018facial}, enable identity theft~\cite{wang2025nullswap,wenger2023sok}, and facilitate unauthorized mass surveillance~\cite{almeida2022ethics}. With billions of facial images accessible online, evaluating the robustness of these systems against advanced evasion techniques has become increasingly critical. Recent studies reveal that deep neural network–based FR models are inherently vulnerable to adversarial perturbations~\cite{fan2020sparse,mirjalili2020privacynet,chhabra2025pridentity,wang2023privacy,le2025diffprivate,wu2018towards}---carefully designed modifications added to otherwise benign images that can mislead recognition systems~\cite{madry2018towards}. These adversarial examples demonstrate strong capabilities for attacking state-of-the-art FR models, even under black-box settings~\cite{komkov2021advhat,jia2022adv}. Consequently, designing advanced adversarial mechanisms for face recognition has emerged as a crucial research direction for biometric security evaluation.

To date, the landscape of adversarial research in face recognition has been studied by three primary methodologies: gradient-based, patch-based, and stealthy-semantic attacks. Early gradient-based methods generate adversarial examples by adding bounded pixel-level perturbations to the input image~\cite{madry2018towards,dong2018boosting}. Although such perturbations may appear visually imperceptible, they often lack robustness under real-world variations such as lighting changes and tend to exhibit limited transferability across models~\cite{nie2022diffusion,qiu2020semanticadv}. Patch-based methods instead synthesize localized adversarial patterns that physically occlude or alter facial regions~\cite{sharif2016accessorize,komkov2021advhat}. While effective in physical environments, these approaches typically introduce conspicuous visual artifacts that compromise stealthiness and are unsuitable for natural image sharing. To improve visual realism, more recent methods attempt to generate semantic adversarial faces by manipulating higher-level facial attributes rather than individual pixels. For example, prior works explored adversarial semantic editing through attribute conditioning or makeup transfer networks~\cite{hu2022protecting}. Although these approaches improve perceptual quality, they often rely on reference images or manually defined semantic cues, which limits scalability and generalization. GAN-based frameworks further enhance realism through learned facial editing operations~\cite{huang2022fakelocator,liu2023gan,huang2020pfa}, but they typically require retraining for new target identities and remain constrained by the generative manifold of GAN models.

Diffusion models have recently emerged as a powerful alternative for high-fidelity image synthesis, offering improved controllability and diversity compared with GAN-based approaches. Leveraging their strong inductive bias for spatial data and perceptually meaningful latent representations~\cite{rombach2022ldm}, diffusion models provide an attractive foundation for generating adversarial facial transformations. Recent studies such as DiffProtect~\cite{liu2023diffprotect}, Adv-Diffusion~\cite{liu2024advdiffusion}, Adv-CPG~\cite{wang2025advcpg}, and Adv-TGD \cite{ahmadieh2026adv} explore diffusion-based adversarial attacks by injecting perturbations into the latent space rather than the raw pixel domain, enabling more stealthy identity manipulation. Despite these advances, existing diffusion-based approaches often rely on iterative optimization or target-specific retraining, which limits efficiency and editing flexibility. Moreover, many approaches rely on rigid heuristic masks, struggling to maintain global photorealistic appearance without introducing localized boundary artifacts.

\begin{figure}[t!]
  \centering
  \includegraphics[width=\linewidth]{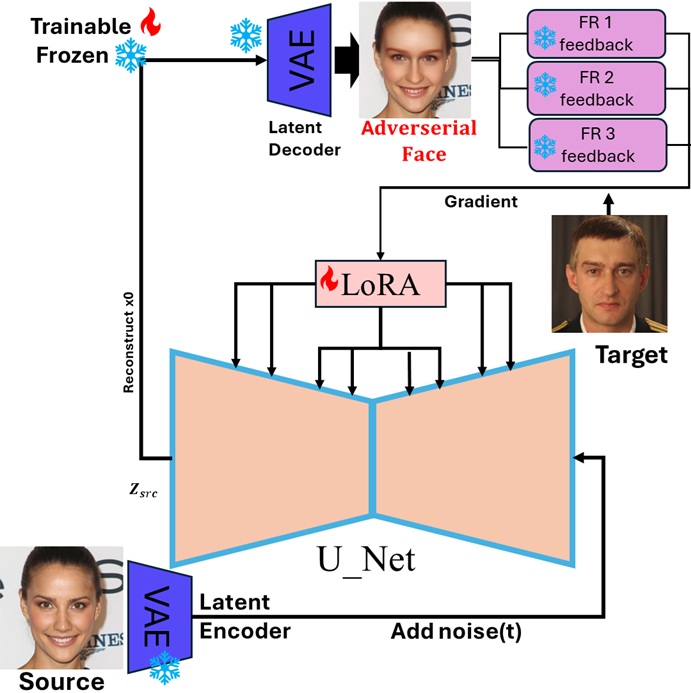} 
  \caption{This architecture diagram illustrates a sophisticated pipeline for creating adversarial faces that mimic one person to a human eye but are misidentified by Face Recognition (FR) systems. The generated image is passed through frozen Face Recognition models to calculate a gradient based on how the FR models perceive the face versus the Target image. That gradient is fed back into the LoRA (Low-Rank Adaptation) weights. It leverages the generative power of Stable Diffusion and the efficiency of LoRA to optimize the output globally. By using LoRA, the system doesn't need to fine-tune the entire massive U-Net. It only trains a few small matrices, making the attack faster and computationally cheaper. Because the optimization is applied to the full image latent, the diffusion model naturally harmonizes the adversarial identity with the original background without relying on heuristic blending or masking.}
  \label{fig:arch}
\end{figure}

To address the limitations of traditional pixel-space adversarial perturbations, we propose \textbf{DiffAttack}, a novel framework that leverages the generative priors of Stable Diffusion to synthesize high-fidelity adversarial faces. Unlike previous methods that rely on computationally expensive global fine-tuning or rigid GAN manifolds, our approach utilizes Low-Rank Adaptation (LoRA) to efficiently optimize the adversarial signal directly within the latent space. As illustrated in Fig.~\ref{fig:arch}, our pipeline begins by encoding a clean source image $x_{src}$ into a latent representation $z_{src}$ using a frozen VAE encoder. To perform the attack, we introduce a fixed amount of Gaussian noise at timestep $t$, producing a noisy latent that serves as the input to the U-Net backbone. The core of our innovation lies in the LoRA-augmented cross-attention layers. Instead of updating the massive U-Net parameters, we inject lightweight, trainable LoRA adapters that are specifically optimized to redirect the diffusion process toward an adversarial identity. The optimization is driven by a multi-source feedback loop. In each iteration, the U-Net predicts the noise, allowing us to reconstruct the clean latent $x_0$. This latent is passed through a frozen VAE decoder to generate the adversarial face in pixel space. To ensure high transferability, we employ gradient feedback mechanism using three state-of-the-art face recognition (FR) models: IR152, IRSE50, and MobileFace. These models calculate the identity distance between the generated adversarial face and a specified target identity. The resulting gradients are backpropagated solely to the LoRA weights, iteratively refining the adversarial features until the FR systems successfully misidentify the source as the target. A critical challenge in adversarial generation is maintaining the visual ``naturalness" of the source image to avoid human detection. Instead of relying on rigid segmentation masks or localized bounding boxes, DiffAttack optimizes the full-image latent representation. By fine-tuning the cross-attention layers globally, the diffusion model naturally harmonizes the adversarial identity features with the original background, lighting, and peripheral attributes. This results in a seamless, photorealistic output that achieves a high attack success rate (ASR) of 84.86\% without the boundary halos common in masked attacks.

Our main contributions are:
\begin{itemize}
\item \textbf{Novel Latent-Space Evasion Framework.} We introduce a unified adversarial pipeline that performs optimization in the latent space rather than the raw pixel space. By utilizing LoRA-augmented cross-attention layers, we enable the automatic learning of effective adversarial semantic appearances. This approach avoids the conspicuous artifacts of traditional gradient-based noise while requiring significantly less computational overhead than global model fine-tuning.

\item \textbf{Black-Box Transferability.} We design a multi-source optimization strategy that leverages white-box feedback from diverse surrogate FR backbones, specifically IR152, IRSE50, and MobileFace. By backpropagating identity-distance gradients directly to lightweight LoRA adapters, our method injects adversarial semantics that generalize beyond the training ensemble. We demonstrate the robustness of this approach through successful transfer-based black-box attacks against the FaceNet system. Notably, DiffAttack achieves high success rates without any exposure to FaceNet's internal architecture or parameters, proving its efficacy against unseen, real-world biometric deployments.

\item \textbf{Global Latent Harmonization.} Instead of relying on rigid spatial masks or localized blending, our framework optimizes the adversarial signal across the entire latent space. This global approach allows the diffusion priors to naturally harmonize the identity shift with the surrounding context (e.g., background and lighting), ensuring strict photorealism and avoiding the boundary artifacts common in localized attacks.

\item \textbf{State-of-the-Art Performance.} We conduct extensive evaluations on the FFHQ and CelebA-HQ datasets. As demonstrated in Table 1, DiffAttack achieves a new state-of-the-art with an average Attack Success Rate (ASR) of 84.86\%, outperforming existing noise-based, makeup-based, and semantic-based methods. Notably, our method achieves near-perfect evasion rates against MobileFace (95.03\%) and IRSE50 (94.24\%) while maintaining superior visual quality.
\end{itemize}  
\section{Related Work}
\label{sec:related_work}

\textbf{Adversarial Attacks on Face Recognition:} Adversarial evasion in face recognition (FR) seeks to induce misclassification through carefully engineered perturbations. Traditional noise-based attacks~\cite{madry2018towards, dong2018boosting} utilize $\ell_p$-bounded pixel noise to deceive models but are frequently plagued by poor transferability to black-box systems and visible "salt-and-pepper" artifacts that trigger human suspicion. To improve stealth, unrestricted attacks~\cite{sharif2019general, zhu2023unrestricted} and makeup-based methods~\cite{yin2021advmakeup, hu2022protecting} embed perturbations within semantic edits. While these improve perceptual quality, they often rely on large-scale makeup datasets that introduce demographic bias or require specific reference images, limiting their generalizability in real-world deployment.

\textbf{Diffusion Models for Image Synthesis:} Diffusion models~\cite{ho2020denoising, song2020score, rombach2022ldm} have redefined the state-of-the-art in generative modeling by iteratively refining Gaussian noise into high-fidelity imagery. The introduction of Latent Diffusion Models (LDM)~\cite{rombach2022ldm} significantly reduced computational costs by operating in a compressed latent space. Recent advancements in parameter-efficient fine-tuning, such as LoRA~\cite{hu2022lora}, allow for the injection of new styles or identities by updating only low-rank matrices within the U-Net's attention layers. While these have been widely used for creative personalization~\cite{gal2022lora, chen2023customdiffusion}, their potential for structured adversarial optimization remains underexplored.

\textbf{Diffusion-Driven Adversarial Evasion:} The integration of diffusion priors into adversarial design offers a promising alternative to raw pixel manipulation. Early attempts like \emph{Adv-Diffusion}~\cite{liu2024advdiffusion} utilized latent perturbations but suffered from high computational latency due to multi-step iterative optimization. Subsequent works like \emph{DiffAM}~\cite{sun2024diffam} improved realism but struggled to maintain high Attack Success Rates (ASR) across diverse FR backbones or relied on rigid, localized heuristic masks that introduce boundary artifacts. Unlike these methods, our proposed \textbf{DiffAttack} leverages full-image, LoRA-based adaptation. By optimizing the adversarial signal globally across the latent space, we allow the diffusion model's inherent generative priors to naturally harmonize the identity shift with the surrounding context, ensuring both high ASR and seamless photorealism.
\section{Methodology}
\label{sec:methodology}

We introduce \textbf{DiffAttack}, a per-sample LoRA fine-tuning framework that synthesizes \emph{targeted} adversarial faces by globally adapting a frozen latent diffusion model with a lightweight, pair-specific adapter. Unlike prior works that rely on heuristic masking or localized blending, our method operates on the entire latent space to ensure natural harmonization. The optimization leverages an ensemble of face recognition models to iteratively guide the LoRA adapters, shifting the identity in the diffusion process while seamlessly maintaining the global structure of the original image.

\subsection{Threat Model}
Adversarial attacks on face recognition (FR) are generally categorized into two distinct scenarios: White-box and Black-box attacks. In a white-box setting, the adversary has full access to the victim model's architecture, internal parameters, and gradients. This allows for the direct optimization of the adversarial loss, which typically yields the highest success rates but is less practical for real-world deployment. Conversely, in a black-box attack, the adversary has no knowledge of the victim system's internal workings. These attacks are further subdivided into query-based methods, where the attacker relies on model output scores, and transfer-based methods, where adversarial examples generated on a surrogate model are used to deceive a different target system.

In this work, we operate under a realistic transfer-based black-box setting for the final victim system, while leveraging a white-box setting during the optimization of our surrogate ensemble. Specifically, we assume the attacker does not have access to the target system $f_{v}$. Instead, we utilize the gradients of a surrogate ensemble consisting of other three models to optimize the adversarial features. The potency of DiffAttack relies on the cross-model transferability of these latent-space perturbations. In our black-box impersonation attack setting, the adversary aims to generate an adversarial image $x_{adv}$ that is visually indistinguishable from a source image $x_{src}$, yet is classified by a victim face recognition (FR) system $f_{v}(\cdot)$ as a target identity $y_{trg}$.

\textbf{Constraints and Adversarial Knowledge:} Following the standard protocol for realistic evasion attacks, we assume the attacker has no access to the target model's internal architecture, weights, or training data. The attack is strictly limited to a query-based or transfer-based setting. Formally, given a source image $x_{src} \in \mathcal{X}$ and a target identity $x_{trg}$, the attacker seeks to find a latent perturbation $\delta$ such that:
\begin{equation}
\min_{\delta} \mathcal{L}_{adv}(f_{v}(x_{adv}), f_{v}(x_{trg})) \quad \text{s.t.} \quad \|x_{adv} - x_{src}\|_p \leq \epsilon
\end{equation}
where $x_{adv} = \text{Dec}(\text{Enc}(x_{src}) + \delta)$ and $f_{v}$ is the black-box target model.

\textbf{Surrogate-Based Optimization:} Since the gradients of the target model $f_{v}$ are inaccessible, we employ a surrogate ensemble $\mathcal{S} = \{f_{s1}, f_{s2}, f_{s3}\}$ to approximate the loss surface of the target system. For example, when attacking the FaceNet model, we leverage the gradients from an ensemble consisting of IR152, IRSE50, and MobileFace. Conversely, to target MobileFace, our surrogate ensemble is composed of FaceNet, IRSE50, and IR152. The adversarial objective is defined by minimizing the distance between the feature embeddings of the adversarial face and the target identity across the surrogate ensemble:
\begin{equation}
\mathcal{L}_{total} = \sum_{f_i \in \mathcal{S}} \lambda_i \cdot \mathcal{D}(f_i(x_{adv}), f_i(x_{trg}))
\end{equation}
where $\mathcal{D}(\cdot)$ denotes a distance metric typically Cosine Similarity and $\lambda_i$ represents the weighting factor for each surrogate model. By optimizing the LoRA weights $\Delta W$ within the diffusion U-Net to minimize $\mathcal{L}_{total}$, we generate adversarial examples that exhibit high transferability to the unseen target model. 

\subsection{Latent Space Parameters}
The attack operates in the latent space $\mathcal{Z}$ provided by a pre-trained VAE.

\textbf{Source Latent ($z_{src}$):} The compressed representation of the source image $x_{src}$, where $z_{src} = \mathcal{E}(x_{src})$. This serves as the structural anchor.

\textbf{Timestep ($t$):} A fixed diffusion step used to add Gaussian noise $\epsilon$ to the latent:
\begin{equation}
z_t = \sqrt{\bar{\alpha}_t} z_{src} + \sqrt{1 - \bar{\alpha}_t} \epsilon
\end{equation}
Selecting an optimal $t$ is crucial; a $t$ too small limits semantic change, while a $t$ too large destroys the source's structural integrity.

\subsection{The LoRA-Augmented U-Net}
Instead of modifying the dense weights $W \in \mathbb{R}^{d \times k}$ of the U-Net, we inject Low-Rank adapters.

\textbf{Rank ($r$):} The bottleneck dimension of the LoRA matrices $A \in \mathbb{R}^{r \times k}$ and $B \in \mathbb{R}^{d \times r}$. We typically set $r \ll \min(d, k)$ (e.g., $r=4$ , $8$, $16$) to ensure efficiency.

\textbf{Updated Weights ($\Delta W$):} The added weight is the product of the two low-rank matrices, scaled by a factor $\alpha/r$:
\begin{equation}
\Delta W = \frac{\alpha}{r} (B \cdot A)
\end{equation}

\textbf{Cross-Attention Layers:} The LoRA weights are specifically applied to the $Q, K, V$ projections within the U-Net's attention blocks, as these layers control the relationship between the visual latent and the identity-conditioning features.

\subsection{Loss Function and Surrogate Parameters}
The optimization aims to minimize a multi-objective loss function $\mathcal{L}_{total}$, which is driven by an ensemble of surrogate models (IR152, IRSE50, MobileFace) denoted as $\mathcal{F}_k$. Our unified identity objective consists of three components:

\textbf{Ensemble Identity Loss ($\mathcal{L}_{id}$):} We push the generated face embedding $f_k(x_{adv})$ toward the target embedding $e_{tgt}^k$ using a cosine similarity hinge, where $\tau_k$ is the verification threshold:
\begin{equation}
\mathcal{L}_{id} = \frac{1}{K} \sum_{k=1}^{K} [\tau_k - \cos(f_k(x_{adv}), e_{tgt}^k)]_+
\end{equation}

\textbf{Directional Guidance ($\mathcal{L}_{dir}$):} To ensure the latent shift follows semantic identity trajectories rather than exploiting superficial shortcuts, we align the observed displacement with the desired source $\to$ target vector:
\begin{equation}
\mathcal{L}_{dir} = \frac{1}{K} \sum_{k=1}^{K} \left(1 - \cos\left( \nu(f_k(x_{adv}) - e_{src}^k), \nu(e_{tgt}^k - e_{src}^k) \right)\right)
\end{equation}
where $\nu(\cdot)$ denotes unit-normalization and $e_{src}^k$ is the source embedding.

\textbf{Source Suppression ($\mathcal{L}_{src}$):} We apply a penalty to discourage the optimizer from drifting back to the original source identity:
\begin{equation}
\mathcal{L}_{src} = \frac{1}{K} \sum_{k=1}^{K} [\cos(f_k(x_{adv}), e_{src}^k) - m]_+
\end{equation}
where $m$ is a small margin tolerance. The total loss is defined as:
\begin{equation}
\mathcal{L}_{total} = \lambda_{id}\mathcal{L}_{id} + \lambda_{dir}\mathcal{L}_{dir} + \lambda_{src}\mathcal{L}_{src}
\end{equation}

\subsection{Evaluation Metrics}
To validate the effectiveness of these parameters, we employed both quality with  four widely used metrics. Attack success rate (ASR), fréchet inception distance (FID), peak signal-to-noise ratio (PSNR), and structural similarity index (SSIM). These metrics is used to assess the attack effectiveness, distribution realism, and visual fidelity of the generated images.

\paragraph{Attack Success Rate (ASR).}
ASR is the primary metric for evaluating evasion effectiveness. In the context of an adversarial attack, ASR quantifies the proportion of adversarial images $x'$ that are successfully identified as the target identity $y_{trg}$ by the victim FR model $f(\cdot)$ at a specific threshold $\tau$. Formally, for $N$ test pairs, ASR is defined as

\begin{equation}ASR = \frac{1}{N} \sum_{i=1}^{N} \mathbf{1}\left(\text{cos}(f(x_i'), f(y_{trg,i})) > \tau\right)\end{equation}where $\mathbf{1}(\cdot)$ is the indicator function and $\text{cos}(\cdot)$ denotes the cosine similarity between feature embeddings. We report ASR at a false acceptance rate (FAR) of 0.01 to ensure high confidence in the attack results. A higher ASR indicates stronger adversarial effectiveness.

\paragraph{Fr\'echet Inception Distance (FID).}
FID evaluates the distribution similarity between adversarial and genuine images using deep features extracted from an Inception network. To ensure that the adversarial faces generated via the latent diffusion process remain within the natural image manifold, we use FID~\cite{heusel2017gans}. FID measures the distance between the distribution of real images $p_r$ and generated adversarial images $p_g$. Using the features from a pre-trained Inception-v3 pool3 layer, FID is defined as:\begin{equation}FID = ||\mu_r - \mu_g||_2^2 + \mathrm{Tr}\left(\Sigma_r + \Sigma_g - 2(\Sigma_r \Sigma_g)^{\frac{1}{2}}\right)\end{equation}Lower FID indicates that DiffAttack generates adversarial samples that are distributionally indistinguishable from authentic facial datasets (FFHQ/CelebA-HQ).

\paragraph{Peak Signal-to-Noise Ratio (PSNR).}
We utilize PSNR to quantify the intensity of the adversarial signal injected into the source image. Since our method operates in the latent space, PSNR serves as a crucial check to ensure that the pixel-level reconstruction does not deviate excessively from the source $x$. It is calculated based on the Mean Squared Error (MSE):\begin{equation}PSNR = 10 \log_{10}\left(\frac{255^2}{\frac{1}{WH} \sum_{i,j} (x_{ij} - x'_{ij})^2}\right)\end{equation}High PSNR values demonstrate that the adversarial perturbations are subtly embedded within the facial structure.

\paragraph{Structural Similarity Index (SSIM).}
Beyond pixel-wise differences, SSIM evaluates the preservation of the source's structural integrity, which is vital for maintaining the user's visual identity to human observers. SSIM evaluates perceptual similarity by comparing luminance ($l$), contrast ($c$), and structure ($s$) information between two images from authentic and adversarial. Given two image patches $x$ and $y$, SSIM is defined as

\begin{equation}SSIM(x,x') = [l(x,x')]^\alpha \cdot [c(x,x')]^\beta \cdot [s(x,x')]^\gamma\end{equation}An SSIM value near $1$ indicates that the DiffAttack successfully preserves the high-frequency structural details (e.g., skin texture and bone structure) of the original subject despite the adversarial identity shift. SSIM values range between $0$ and $1$, with values closer to $1$ indicating higher structural similarity.

\subsection{Global Generative Harmonization}
\label{sec:harmonization}
Because our framework avoids the use of face-local bounding boxes, heuristic masking, or pixel-space segmentation, we eliminate the need for traditional image blending techniques such as Poisson seamless cloning. Instead, DiffAttack relies entirely on the inherent generative priors of the diffusion model to integrate the adversarial identity shift.

By operating on the full-image latent $z_{src}$ and utilizing a carefully tuned noise anchor at timestep $t$, the global structure, lighting, and background of the original image are preserved. As the gradients from the FR models update the LoRA adapters to alter the facial identity features, the U-Net simultaneously reconstructs the surrounding non-facial regions. This global optimization naturally hallucinates a smooth, seamless transition between the adversarially altered face and the unperturbed background, yielding a photorealistic final output free from boundary halos or artificial edge artifacts.
\section{Experiments}
\label{sec:experiments}

We evaluate the effectiveness of our method in \emph{black-box} attack settings against robust face recognition (FR) models and compare with recent state-of-the-art.

\subsection{Experimental Setting}

\begin{table*}[t]
\centering
\caption{Attack Success Rate (ASR\%) on \textbf{FFHQ} and \textbf{CelebA-HQ}}
\label{tab:asr_results_columnar}
\resizebox{\linewidth}{!}{%
\begin{tabular}{ll|cccc|cccc|c}
\toprule
& & \multicolumn{4}{c|}{\textbf{FFHQ}} & \multicolumn{4}{c|}{\textbf{CelebA-HQ}} & \\
\cmidrule(lr){3-6} \cmidrule(lr){7-10}
\textbf{Category} & \textbf{Method} &
\textbf{IR152} & \textbf{IRSE50} & \textbf{FaceNet} & \textbf{Mobile} &
\textbf{IR152} & \textbf{IRSE50} & \textbf{FaceNet} & \textbf{Mobile} &
\textbf{Avg.} \\
\midrule

& Clean
& 3.20 & 2.20 & 2.10 & 4.90
& 2.10 & 5.61 & 0.80 & 13.60
& 4.31 \\
\midrule

\multirow{3}{*}{Noise-Based}
& TI\mbox{-}DIM~\cite{dong2019translationinvariant}
& 42.26 & 63.91 & 14.28 & 51.71
& 34.08 & 60.51 & 13.27 & 51.25
& 41.41 \\

& TIP\mbox{-}IM~\cite{yang2021towards}
& 44.86 & 65.36 & 58.03 & 50.47
& 40.02 & 55.57 & 37.90 & 48.07
& 50.04 \\

& P3\mbox{-}Mask~\cite{chow2024personalized}
& 70.98 & 82.79 & 56.18 & 68.57
& 71.27 & 80.90 & 58.43 & 67.55
& 69.58 \\
\midrule

\multirow{4}{*}{\shortstack[l]{Makeup /\\Attribute}}
& CLIP2Protect~\cite{shamshad2023clip2protect}
& 50.56 & 83.93 & 43.66 & 74.00
& 46.20 & 78.53 & 41.29 & 71.43
& 61.20 \\

& DFPP~\cite{shamshad2024makeup}
& 52.62 & 87.91 & 50.57 & 77.69
& 45.00 & 78.37 & 44.01 & 69.97
& 63.24 \\

& DiffAM~\cite{sun2024diffam}
& 65.22 & 87.59 & \underline{63.01} & 86.87
& 62.14 & 86.97 & 61.10 & 81.97
& 74.49 \\

& DiffAIM~\cite{wang2025diffusion}
& \underline{82.20} & \underline{91.04} & \textbf{68.21} & \underline{94.21}
& \underline{80.56} & \underline{90.37} & \textbf{64.14} & \underline{93.10}
& \underline{82.85} \\
\midrule

\multirow{4}{*}{\shortstack[l]{Portrait /\\Semantic}}
& Adv\mbox{-}Diffusion~\cite{liu2024advdiffusion}
& 49.40 & 79.31 & 29.91 & 65.49
& 51.25 & 79.22 & 33.90 & 68.66
& 57.14 \\

& TCA$^2$~\cite{tca2025aaai}
& 53.63 & 77.31 & 41.62 & 72.21
& 52.50 & 76.21 & 40.72 & 71.92
& 60.76 \\

& Adv-CPG~\cite{wang2025advcpg}
& 75.26 & 91.03 & 62.84 & 89.94
& 76.96 & 88.72 & \underline{63.50} & 87.95
& 79.65 \\

\rowcolor{gray!15}
& \textbf{DiffAttack (Ours)}
& \textbf{87.72}
& \textbf{93.45}
& 62.63
& \textbf{94.24}
& \textbf{89.50}
& \textbf{93.65}
& 62.63
& \textbf{95.03}
& \textbf{84.86} \\

\bottomrule
\end{tabular}%
}
\end{table*}

\paragraph{Dataset:} We evaluate DiffAttack using two widely adopted, high-quality facial image datasets CelebA-HQ~\cite{karras2017progressive} and FFHQ~\cite{karras2019style}. In line with the protocol described in~\cite{wang2025advcpg, liu2024advdiffusion}, a subset of 1,000 face images representing distinct identities is selected from each dataset along with five more images as a target image. 1,000 face images are divided into five groups, with five target images randomly chosen per group. As a result, each dataset contains five groups, each comprising 200 source–target image pairs.

\paragraph{Implementation details:}
We build on Stable Diffusion v2.1 at $768{\times}768$. For each $(I_{\text{src}}, I_{\text{tgt}})$ pair we fine-tune a \emph{pair-specific} LoRA adapter injected into U-Net cross-attention projections \texttt{to\_q}, \texttt{to\_k}, \texttt{to\_v}, \texttt{to\_out.0}. We use LoRA rank $r{=}16$ and $\alpha{=}64$ with dropout $0.08$. Training runs for $45$ steps (AdamW, base LR $3{\times}10^{-5}$ with late linear decay to $1{\times}10^{-5}$). We use a single denoising slice per step with a fixed time index fraction $\tau{=}0.6$ and a DDPM scheduler. Re-evaluation scores are computed directly on the generated outputs, without any heuristic masking or post-hoc blending back into the original frame.

\paragraph{Robustness across target identities.}
Generating adversarial examples for arbitrary target faces is crucial in practice. To assess how sensitive DiffAttack is to the choice of target identity, we randomly select several target images from CelebA-HQ and FFHQ and form multiple groups of source--target pairs. DiffAttack consistently produces high-fidelity adversarial faces that successfully mimic diverse targets. The overall structure and background remain stable across different targets, indicating that our method is robust to changes in target identity rather than overfitting to a specific face.

\paragraph{Global vs. Localized Optimization.}
A key design choice in DiffAttack is the use of global, full-image latent optimization rather than relying on heuristic masks or bounding boxes. Our experiments show that while localized masking restricts changes strictly to the face, it frequently introduces unnatural boundary halos and gradients that fail under human scrutiny or simple artifact detection. In contrast, by fine-tuning the LoRA adapters across the entire spatial latent, DiffAttack leverages the diffusion model's natural generative priors to seamlessly harmonize the adversarial identity shift with the surrounding background and lighting, leading to highly photorealistic and stealthy evasion.

\paragraph{Identity loss components.}
Our identity objective combines an ensemble-based hinge term with a directional source $\!\to\!$ target component and source-suppression. We compare the full loss with variants that progressively remove these components. When using only a basic hinge loss, we observe that some adversarial examples partially drift between source and target identities, especially under challenging poses. Adding the directional term encourages a more consistent movement of embeddings toward the target manifold, while explicit source-suppression helps prevent residual similarity to the original identity. 

\paragraph{Evaluation Metrics:}
We evaluate our approach by independently assessing attack performance and image quality.
To measure attack performance, we adopt the Attack Success Rate (ASR)~\cite{jia2022adv, liu2024advdiffusion, wang2025advcpg}, which quantifies the proportion of adversarial examples that successfully deceive the target model. 
\begin{equation}
\text{ASR} = \frac{1}{K} \sum_{I} \mathbb{I}\big( \cos(h(I_t), h(I_{adv})) > \tau \big) \times 100\%,
\end{equation}
where $\mathbb{I}$ denotes the indicator function, $K$ represents the total number of face images, and $\tau$ is the decision threshold. 
$I_t$ and $I_{adv}$ correspond to the target and adversarial face images, respectively. 
The threshold $\tau$ is set to yield a 0.01 False Acceptance Rate (FAR) for each victim model, following the same configuration~\cite{jia2022adv, liu2024advdiffusion, wang2025advcpg}. Additional details are provided in the supplementary material.
For image quality assessment, we employ three standard metrics: Frechet Inception Distance (FID), Peak Signal-to-Noise Ratio (PSNR), and Structural Similarity Index (SSIM). These metrics collectively evaluate the perceptual similarity and visual fidelity of the generated adversarial images.
\subsection{Comparison with SOTA Methods}

\paragraph{Quantitative results.}
Table~\ref{tab:asr_results_columnar} summarizes the attack success rates (ASR) under the black-box setting; underlined entries denote the highest value in each column. For face verification, we set the decision threshold $\tau$ per model at $\mathrm{FAR}=0.01$. Our method achieves strong black-box transferability, particularly on IR152, IRSE50, and MobileFace, and substantially outperforms representative noise-, makeup-, and semantic-based baselines.

\paragraph{Image Quality Assessment}
To further validate the imperceptibility of the proposed adversarial attack, we evaluate the visual quality of adversarial examples compared with representative SOTA baselines. Following common practice, we adopt Frechet Inception Distance (FID)~\cite{heusel2017gans}, Peak Signal-to-Noise Ratio (PSNR), and Structural Similarity Index (SSIM) as evaluation metrics, where a lower FID and higher PSNR/SSIM indicate better visual fidelity. Table~\ref{tab:quality_results} reports the quantitative results aggregated across multiple runs. Among all competitors, our method achieves the highest SSIM (0.961) and a competitive FID of 27.58, indicating strong perceptual quality and minimal visual distortion. Compared with DiffAM and Adv-Diffusion, DiffAttack produces adversarial images with more consistent facial structure and texture realism while preserving semantic integrity. These results demonstrate that our approach generates natural and visually indistinguishable adversarial examples, achieving an effective balance between attack performance and perceptual fidelity. 

As can be seen in Fig.~\ref{fig:teaser}, the comparison of the DiffAttack method against current State-of-the-Art (SOTA) benchmarks reveals a significant evolution in how adversarial face swaps are executed under black-box constraints using the Face++ API. Traditional Noise-Based approaches, such as TIP-IM, achieve relatively high identity alignment scores reaching up to 75.7 in the first row, but they suffer from severe visual degradation. These methods introduce ``ghosting" artifacts and unnatural skin textures that make the manipulation obvious to a human observer, despite successfully deceiving the facial recognition system. In contrast, Makeup-Based methods like DiffAM and DiffAIM attempt to mask the adversarial intent behind cosmetic changes. However, these often result in inconsistent performance and aesthetic failures. For instance, DiffAM produces the lowest identity alignment score in the first sample (54.3) and applies heavy, saturated purple tones that look highly artificial. While DiffAIM improves upon this with smoother blending and higher scores (e.g., 72.0), it still lacks the structural precision found in more advanced diffusion-based frameworks. The most sophisticated competition arises from Facial Semantic methods like Adv-Diffusion. While Adv-Diffusion demonstrates a high capacity for identity shifting peaking at a score of 78.4 in the third row, it frequently introduces structural instability. Noticeable ``leakage" occurs in the hair and background, and the facial proportions can become slightly warped. DiffAttack (Ours) distinguishes itself by achieving a near-perfect equilibrium between machine-targeted deception and human-perceived photorealism. It maintains high Face++ confidence scores, such as 76.6 and 76.1, while perfectly preserving the original source's texture, lighting environment, and subtle expressions. This ensures that the generated image is not only identified as the target by the Open API but remains indistinguishable from a genuine photograph to the human eye, representing a superior advancement in black-box attack stealth.

\begin{table}[h!]
\centering
\caption{Perceptual quality comparison. Lower is better for FID; higher is better for PSNR/SSIM.}
\label{tab:quality_results}
\begin{tabular}{lccc}
\toprule
\textbf{Method} & \textbf{FID} $\downarrow$ & \textbf{PSNR} $\uparrow$ & \textbf{SSIM} $\uparrow$ \\
\midrule
TIP-IM & 38.73 & 33.21 & 0.924 \\
Adv-Diffusion & 22.58 & 28.85 & 0.805 \\
DiffAM & 26.10 & 20.53 & 0.886 \\
Adv-CPG & 26.07 & 29.99 & 0.897 \\
DiffAIM & 23.23 & 25.39 & 0.739 \\
\textbf{DiffAttack (Ours)} & 27.58 & 24.54 & 0.961 \\
\bottomrule
\end{tabular}
\end{table}

\section{Ethical Considerations}
Our research explores vulnerabilities in face recognition systems using generative AI. While the proposed Adv-TGD framework can synthesize impersonation images, we emphasize that this work is intended for defensive security analysis and privacy protection. All experiments were conducted using publicly available research datasets (CelebA-HQ, FFHQ, LADN). We do not release any pre-trained LoRA weights for specific individuals, and our methodology is designed to inform the development of more robust biometric verification systems against generative threats.
\section{Conclusion}
\label{sec:conclusion}

In this paper, we introduced \textbf{DiffAttack}, a global latent-space optimization framework for generating high-fidelity adversarial faces using a frozen Stable Diffusion~2.1 backbone. By leveraging full-image LoRA fine-tuning and a unified identity objective comprising ensemble hinges, directional guidance, and source suppression \textbf{DiffAttack} enables targeted identity manipulation that is both parameter efficient and structurally consistent. Unlike traditional pixel-space adversarial noise, our method operates globally in the generative latent space, naturally harmonizing the adversarial identity shift with the surrounding unperturbed context. Extensive experiments on CelebA-HQ and FFHQ under a strict black-box setting demonstrate that \textbf{DiffAttack} achieves exceptional transferability across industry standard FR models. Our method attains a SOTA mean ASR of \textbf{84.86}\%, outperforming representative noise-based, makeup-based, and semantic baselines. Crucially, by avoiding rigid heuristic masks, \textbf{DiffAttack} maintains strong perceptual quality, achieving a PSNR of \textbf{24.54}\,dB and an SSIM of \textbf{0.961}, ensuring that the adversarial faces remain photorealistic and free from localized boundary artifacts.

\section{Acknowledgements}
This project received partial support from the National Science Foundation through Grants No. 2603113.

{\small
\bibliographystyle{IEEEtran}
\bibliography{main}
}

\end{document}